\documentclass[11pt]{article}

\usepackage[preprint]{acl}

\usepackage{times}
\usepackage{latexsym}

\usepackage[T1]{fontenc}

\usepackage[utf8]{inputenc}
\usepackage{amsmath}
\usepackage{microtype}

\usepackage{inconsolata}

\usepackage{graphicx}
\usepackage{booktabs}
\usepackage{float}
\usepackage{multirow}
\usepackage{tabularx}
\usepackage{booktabs}
\usepackage{url}
\definecolor{myred}{RGB}{252,152,114}
\definecolor{mygreen}{RGB}{102,194,165}
\usepackage[most]{tcolorbox}
\usepackage[table]{xcolor}
\definecolor{PromptFrame}{HTML}{CBD5E1}
\definecolor{PromptHeader}{HTML}{5e6c7d}
\definecolor{SystemBg}{HTML}{EFF6FF}
\definecolor{SystemTitle}{HTML}{53b8e7}
\definecolor{ContextBg}{HTML}{F8FAFC}
\definecolor{ContextTitle}{HTML}{475569}

\newtcolorbox{prompttitlebox}[1]{
  enhanced,
  colback=PromptHeader,
  colframe=PromptHeader,
  boxrule=0pt,
  arc=1mm,
  left=2mm,
  right=2mm,
  top=1mm,
  bottom=1mm,
  before skip=8pt,
  after skip=0pt,
  fontupper=\bfseries\color{white},
}

\newtcblisting{systemprompt}{
  enhanced,
  breakable,
  listing only,
  listing options={
    basicstyle=\small\ttfamily,
    breaklines=true,
    columns=fullflexible,
    keepspaces=true,
    showstringspaces=false
  },
  colback=SystemBg,
  colframe=PromptFrame,
  boxrule=0.4pt,
  arc=1mm,
  left=3mm,
  right=3mm,
  top=2mm,
  bottom=2mm,
  title=System Prompt,
  colbacktitle=SystemTitle,
  coltitle=white,
  fonttitle=\small\bfseries\sffamily,
  toptitle=1mm,
  bottomtitle=1mm,
  before skip=0pt,
  after skip=3pt
}

\newtcblisting{contextprompt}{
  enhanced,
  breakable,
  listing only,
  listing options={
    basicstyle=\small\ttfamily,
    breaklines=true,
    columns=fullflexible,
    keepspaces=true,
    showstringspaces=false
  },
  colback=ContextBg,
  colframe=PromptFrame,
  boxrule=0.4pt,
  arc=1mm,
  left=3mm,
  right=3mm,
  top=2mm,
  bottom=2mm,
  title=Context Prompt,
  colbacktitle=ContextTitle,
  coltitle=white,
  fonttitle=\small\bfseries\sffamily,
  toptitle=1mm,
  bottomtitle=1mm,
  before skip=0pt,
  after skip=6pt
}

\usepackage{amsmath}

\newcommand{\calF}{\mathcal{F}}
\newcommand{\calT}{\mathcal{T}}

\newcommand{\calD}{\mathcal{D}}

\usepackage{amsmath,amsfonts,bm}

\def\eqref#1{equation~\ref{#1}}

\def\1{\bm{1}}

\def\ve{{\bm{e}}}
\def\vf{{\bm{f}}}

\def\vn{{\bm{n}}}

\def\vp{{\bm{p}}}

\def\vy{{\bm{y}}}

\DeclareMathAlphabet{\mathsfit}{\encodingdefault}{\sfdefault}{m}{sl}
\SetMathAlphabet{\mathsfit}{bold}{\encodingdefault}{\sfdefault}{bx}{n}

\title{ExTax: Explainable Disinformation Detection via Persuasion, Emotion, and
Narrative Role Taxonomies}

\author{
 \textbf{Shang Luo\textsuperscript{1}},
 \textbf{Yingguang Yang\textsuperscript{2}},
 \textbf{Zhenchen Sun\textsuperscript{3}},
 \textbf{Yang Liu\textsuperscript{4}},
\\
 \textbf{Bin Chong\textsuperscript{1}},
 \textbf{Jingru Chen\textsuperscript{5}},
 \textbf{Yancheng Chen\textsuperscript{6}},
 \textbf{Jiayu Liang \textsuperscript{7}},
\\
 \textbf{Kefu Xu\textsuperscript{1}},
 \textbf{Hao Peng\textsuperscript{8}},
 \textbf{Philip S. Yu\textsuperscript{9}},
\\
 \textsuperscript{1}Peking University, 
 \textsuperscript{2}University of Science and Technology of China, \\
 \textsuperscript{3}North China University of Science and Technology,
 \textsuperscript{4}Tsinghua University, \\
 \textsuperscript{5}Nanjing University of Aeronautics and Astronautics, 
 \textsuperscript{6}University of Chinese Academy of Sciences,  \\
 \textsuperscript{7}Soochow University, 
 \textsuperscript{8}Beihang University,
 \textsuperscript{9}University of Illinois Chicago
\\
 \small{
   \textbf{Correspondence:} \href{chongbin@pku.edu.cn}{chongbin@pku.edu.cn}
 }
}

\begin{document}
\maketitle
\begin{abstract}
The democratization of LLMs has accelerated the generation and circulation of highly fluent disinformation, making traditional syntax-semantic verification increasingly insufficient. Such deception rarely relies solely on surface-level falsity; instead, it often combines persuasive rhetoric, emotional manipulation, and narrative role construction to influence readers' interpretations through multiple cognitive pathways. However, existing detectors typically emphasize isolated signals---such as syntax, external knowledge, persuasion, or affective cues---and therefore struggle to capture the multi-faceted manipulative intents underlying disinformation or provide human-auditable explanations. To address this gap, we present \textbf{ExTax}, a taxonomy-aligned framework for explainable disinformation detection. ExTax unifies persuasive rhetoric, emotional manipulation, and narrative roles into a 17-dimensional taxonomic space, covering 6 persuasive-rhetoric strategies, 5 emotional-manipulation methods, and 6 narrative-role categories. It elicits attributes from multiple frontier LLMs, reconciles their disagreements through Entropy-driven Dynamic Label Smoothing, and fuses the resulting taxonomic representations with contextual encodings via Heterogeneous Multi-Head Attention, grounding each prediction in an interpretable manipulation profile. Across five cross-domain and cross-genre benchmarks, ExTax achieves an overall Macro $F_1$ of $0.8456$, outperforming state-of-the-art deep learning and LLM-based baselines. It also remains robust under severe genre imbalance, where the strongest deep baseline degrades from $0.9454$ to $0.6194$. Our code is available at \url{https://anonymous.4open.science/r/ExTax-1FBD/}.
\end{abstract}

\section{Introduction}
\label{sec:introduction}
Disinformation detection has become increasingly challenging as social media platforms and generative AI systems accelerate the production and circulation of deceptive yet highly fluent content. Such content rarely operates through surface wording or outright falsity alone; instead, it often combines rhetorical strategies, emotional manipulation, and narrative framing to attack or simplify targets, evoke high-arousal emotions, and assign actors to morally charged roles. In this way, disinformation shapes public interpretation and sharing behavior through multiple cognitive routes \citep{Saeed2022CrowdsourcedFA, lothIndustrializedDeceptionCollateral2026}. Meanwhile, alarmist discourse around disinformation may itself reshape public risk perception, making reliable and interpretable detection increasingly important \citep{Jungherr2024NegativeDE}.

Recent studies improve robustness by incorporating richer textual, contextual, or external signals. Syntax-aware models capture dependency-level distortions \citep{Xiao2024MSynFDMS}, dynamic representation methods adapt feature views to instance-level difficulty \citep{Farhangian2025DRESFN}, and knowledge- or retrieval-augmented LLM frameworks incorporate external evidence or affective cues to enable more timely detection \citep{jinDynamicKnowledgeUpdateDriven2025, liuRAEmoLLMRetrievalAugmented2025}. Intent-aware detection additionally suggests that communicative purposes provide stable signals beyond surface wording \citep{Wang2025BridgingTA}. However, most existing methods still model syntax, knowledge, emotion, persuasion, or intent in isolation, leaving underexplored how multiple human-interpretable dimensions jointly reveal the manipulative structure of disinformation.

We argue that robust disinformation detection should reason over three complementary facets of manipulation: persuasive rhetoric, emotional manipulation, and narrative roles. These facets characterize which rhetorical strategy is employed, which emotional-manipulation method is activated, and which morally charged role is assigned. Grounded in persuasion-oriented taxonomies, emotional manipulation studies, as well as narrative framing research \citep{piskorskiMultilingualMultifacetedUnderstanding2023, jamiesonFlaggingEmotionalManipulation2025, zhaoHowIndividualsCope2024, Mahmoud2025EntityFA}, these facets provide human-interpretable manipulation evidence that extends beyond single-cue reasoning.

Realizing such a unified taxonomy introduces two challenges: no existing corpus jointly annotates these facets, and attributes elicited from a single LLM may inherit model-specific biases. To address these issues, we propose \textbf{ExTax}, an \textbf{Ex}plainable \textbf{Tax}onomy-aligned framework for disinformation detection. ExTax elicits taxonomy-guided attributes from multiple frontier LLMs, regularizes their disagreements through entropy-driven dynamic label smoothing, and constructs a unified 17-dimensional taxonomic representation consisting of 6 persuasive-rhetoric strategies, 5 emotional-manipulation methods, and 6 narrative-role categories. A taxonomy-aligned heterogeneous attention module then integrates this representation with contextual encodings, grounding each prediction in a per-sample manipulation profile.

Across five cross-domain, cross-genre benchmarks, ExTax achieves a Macro $F_1$ of $0.8456$, outperforming the strongest LLM and deep-learning baselines by +2.0 and +3.1 points, respectively. It also remains robust under severe genre imbalance, where the strongest deep baseline degrades from $0.9454$ to $0.6194$ across genres.

Overall, ExTax makes three key contributions: (1) \textbf{Multifaceted taxonomic modeling.} 
    We reframe disinformation detection as reasoning over three complementary facets of manipulation---persuasive rhetoric, emotional manipulation, and narrative roles---and instantiate these facets as a unified 17-dimensional taxonomic representation. 
    (2) \textbf{Noise-robust attribute synthesis.} 
    We aggregate taxonomy-guided outputs from multiple frontier LLMs and regularize noisy hard decisions through entropy-driven dynamic label smoothing, thereby transforming inter-LLM disagreement into reliable synthetic supervision. 
    (3) \textbf{Taxonomy-aligned detection and explanation.} 
    We integrate learned taxonomic signals with contextual encodings through heterogeneous attention, enabling robust disinformation detection while exposing rhetorical, affective-manipulative, and narrative evidence underlying each prediction.

\section{Related Work}
\label{sec:related_work}
\paragraph{Disinformation detection.}
Disinformation detection has been studied under both supervised learning and LLM-based prompting paradigms. Supervised approaches typically rely on lexical, semantic, engagement, or contextual features to learn task-specific detectors \citep{aslamFakeDetectDeep2021}, while recent zero- and few-shot LLM methods address data scarcity and cross-domain generalization \citep{pelrine-etal-2023-towards, bangMultitaskMultilingualMultimodal2023}. Prompting strategies such as zero-shot Chain-of-Thought \citep{lucasFightingFireFire2023} substantially improve detection performance. However, content-level methods may still overfit to dataset artifacts and generalize poorly across genres and topics, motivating supervised detectors that integrate structured, human-interpretable stylistic, rhetorical, or contextual evidence \citep{modzelewskiPCoTPersuasionAugmentedChain2025}.

\paragraph{Emotion, persuasion, and narrative framing in disinformation.}
Communication research has established emotion as central to how audiences process misleading content. The Appraisal-Tendency Framework \citep{hanFeelingsConsumerDecision2007} explains how discrete emotions such as anger and sadness produce distinct cognitive appraisals. \citet{zhaoHowIndividualsCope2024} extended this framework to narrative disinformation, showing that anger prompts heuristic action while sadness fosters deliberation, and \citet{paletzEmotionalContentSharing2023} found that discrete emotions better predict post virality than valence or arousal. \citet{bagoEmotionMayPredict2022} further showed that emotional reliance increases susceptibility to fake news. In parallel, NLP research has produced persuasive-technique taxonomies \citep{piskorskiSemEval2023TaskDetecting2023}, anchoring shared tasks on propaganda and manipulation detection. Entity framing and role portrayal provide a complementary lens for analyzing how actors are positioned within news narratives \citep{Mahmoud2025EntityFA}.

\paragraph{Integrating high-level communicative signals.}
Recent work integrates such signals into detection pipelines. RAEmoLLM \citep{liuRAEmoLLMRetrievalAugmented2025} leverages emotional embeddings for cross-domain disinformation detection, while PCoT \citep{modzelewskiPCoTPersuasionAugmentedChain2025} infuses persuasion knowledge into LLM reasoning. \citet{kamaliUsingPersuasiveWriting2024} further showed that persuasive writing strategies can serve as intermediate signals for health disinformation detection. Unlike prior work that treats emotion, persuasion, and narrative framing as separate evidence sources, ExTax jointly models them within a unified 17-dimensional taxonomic space and mitigates noise in LLM-elicited attributes through entropy-driven soft labeling.

\section{Preliminaries}
\label{sec:preliminaries}
\subsection{Problem Formulation}
Disinformation detection is defined as the task of identifying false, inaccurate, or misleading information that is intentionally designed, presented, and promoted to cause public harm or generate profit. Formally, this task is structured as a binary classification problem in which a model $\calF$ evaluates an input text $T$, such as a news article or a social media post, to determine its credibility. Given a dataset $\calD = \{(T_i, y_i)\}_{i=1}^n$ where $T_i$ represents the content and $y_i \in \{0, 1\}$ is the ground-truth label, the objective is to learn a predictive function $\vf_{det}: \calT \rightarrow \{0, 1\}$ such that:
\begin{equation}
    \label{eq:problem}
    \hat{y}_i = \vf_{det}(T_i),
\end{equation}
where $\hat{y}_i = 1$ denotes that the text $T_i$ is classified as disinformation, and $\hat{y}_i = 0$ indicates it is reliable information.

\subsection{Taxonomy of Disinformation: Persuasion, Emotion, and Narrative Roles}
ExTax operationalizes the three-route cognitive triangulation introduced in §~\ref{sec:introduction} through three complementary taxonomic dimensions, each grounded in a distinct strand of prior literature. For \textbf{Persuasion}, we adopt the hierarchical taxonomy of \citet{piskorskiMultilingualMultifacetedUnderstanding2023}, which organizes 23 fine-grained persuasive techniques into six high-level strategies: Attack on Reputation, Justification, Simplification, Distraction, Call, and Manipulative Wording. For \textbf{Emotion}, we consider five affective states associated with manipulative content---fear, anger, hope, anxiety, and sadness---motivated on prior findings on emotional manipulation and narrative misinformation \citep{jamiesonFlaggingEmotionalManipulation2025,zhaoHowIndividualsCope2024}. For \textbf{Narrative Roles}, we build upon the entity-framing taxonomy of \citet{Mahmoud2025EntityFA} and adopt six role categories: Ethical Stabilizers, Altruistic Catalysts, Overt Aggressors, Deceptive Subversives, Institutional Toxins, and Marginalized Sufferers. Together, these three dimensions constitute a unified 17-dimensional taxonomic space. Detailed category definitions are provided in Appendix~\ref{app:taxonomies}.

\section{Methods}
\begin{figure*}[t]
\centering
\includegraphics[width=\textwidth]{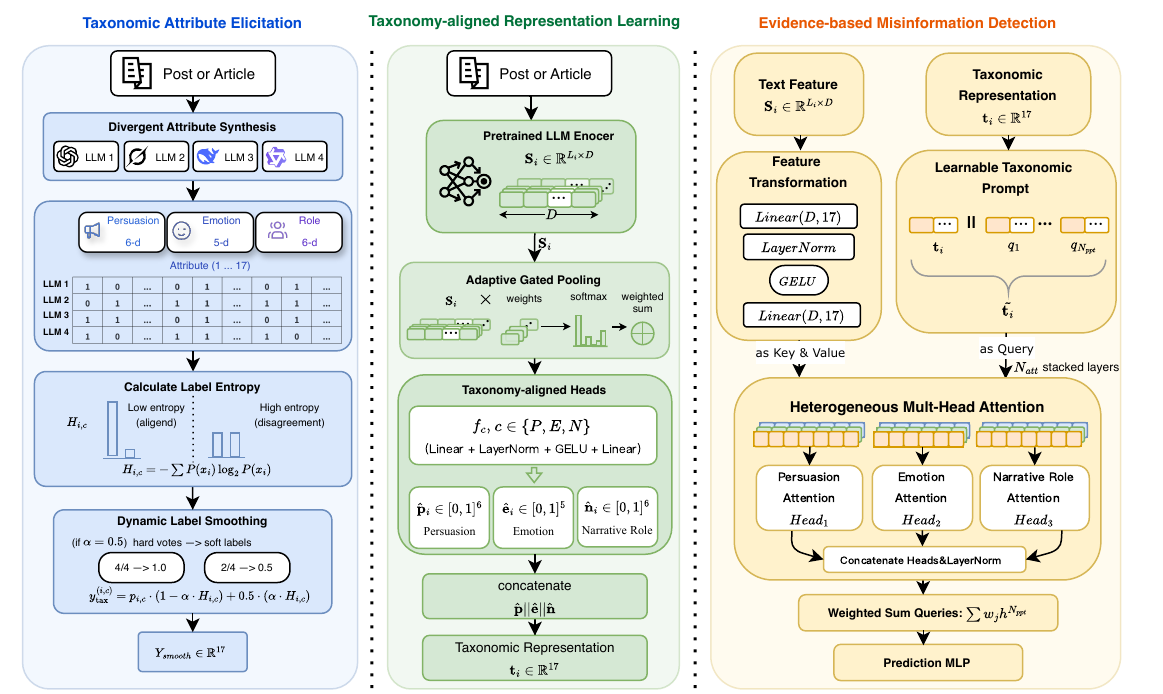}
\caption{Overall framework of ExTax.}
\label{fig:architecture}
\end{figure*}
The architecture of ExTax is illustrated in Figure~\ref{fig:architecture}, and comprises three core components: (1) Taxonomic Attribute Elicitation, which employs multiple LLMs to extract Persuasion, Emotion, and Narrative Role attributes and applies entropy-driven dynamic label smoothing to mitigate model noise; (2) Taxonomy-aligned Representation Learning, which aligns posts or news articles with a unified 17-dimensional taxonomic space; and (3) Taxonomy-based Disinformation Detection, which integrates taxonomic attributes with contextual text encodings for final prediction. Additional architectural details are provided in Appendix~\ref{app:method}.

\subsection{Taxonomic Attribute Elicitation}

Since no large-scale corpus jointly labels the three facets of manipulation, we construct soft taxonomic supervision by querying $N_{LLM} = 4$ top-tier LLMs and reconciling their disagreement. Specifically, for each input $T_i \in \mathcal{D}$, the Taxonomic Attribute Elicitation process extracts attributes associated with Persuasive Rhetoric, Emotional Manipulation, and Narrative Roles. This process yields an augmented dataset $\mathcal{D}_{\text{tax}} = \{(T_i, y_i, \vy_{\text{tax}}^{(i)})\}_{i=1}^n$, where $\vy_{\text{tax}}^{(i)} \in \mathcal{Y}_{\text{smooth}}$ is a 17-dimensional dynamically smoothed soft target vector encoding the refined presence probability of each taxonomic category in $T_i$.

\paragraph{Divergent attribute synthesis.} For each text $T_i$, we query four frontier LLMs $\{M_k\}_{k=1}^{4}$ with taxonomy-specific prompts (Appendix~\ref{app:prompts}) to obtain three binary feature vectors over the persuasive-rhetoric $(|\mathcal{C}_P| = 6)$, emotional-manipulation $(|\mathcal{C}_E| = 5)$, and narrative-role $(|\mathcal{C}_N| = 6)$ category sets:
$\vp_{i,k} \in \{0,1\}^{6},
\ve_{i,k} \in \{0,1\}^{5},
\vn_{i,k} \in \{0,1\}^{6}$.
A component is assigned a value of 1 if the LLM identifies the corresponding category in $T_i$. The raw outputs collected from all four models form the raw taxonomic dataset $\mathcal{R}$, on which we subsequently apply the entropy-driven smoothing procedure described below.

\paragraph{Entropy-driven dynamic label smoothing.}
To reduce noise in LLM-elicited attributes, we transform the hard decisions produced by four LLMs into entropy-aware soft labels. For notational simplicity, we use $z_{i,k}^{(c)}$ to denote the binary output of the $k$-th LLM for any category $c \in \mathcal{C}_P \cup \mathcal{C}_E \cup \mathcal{C}_N$. For each sample $T_i$ and category $c$, we compute the positive vote proportion:
\begin{equation}
    \label{eq:decision}
    p_{i,c} = \frac{1}{4} \sum_{k=1}^{4} z_{i,k}^{(c)},
\end{equation}
where $p_{i,c}$ represents the fraction of LLMs predicting the presence of category $c$. We then estimate the inter-model uncertainty using binary entropy:
\begin{equation}
\label{eq:entropy}
H_{i,c} = - \left[
p_{i,c}\log_2 p_{i,c} + (1-p_{i,c})\log_2(1-p_{i,c})
\right],
\end{equation}
with $0\log_2 0=0$. The final smoothed taxonomic target is defined as:
\begin{equation}
    \label{eq:tax_target}
    y_{\text{tax}}^{(i,c)}
    = p_{i,c}(1-\alpha H_{i,c}) + 0.5(\alpha H_{i,c}),
\end{equation}
where $\alpha$ controls the smoothing strength. Thus, high-disagreement labels are pulled toward the neutral value $0.5$, while high-agreement labels remain close to the original vote proportion. This procedure is applied to all categories in $\mathcal{C}_P \cup \mathcal{C}_E \cup \mathcal{C}_N$, yielding the smoothed taxonomic targets $\mathcal{Y}_{\text{smooth}}$.

\subsection{Taxonomy-aligned Representation Learning} 
The objective of the Taxonomy-aligned Representation task is to map news articles into a multi-faceted taxonomic space defined by Persuasion, Emotion, and Narrative Roles. Specifically, given the textual dataset $\mathcal{D}$ and the dynamically smoothed taxonomic labels $\mathcal{Y}_{\text{smooth}}$, the task requires the model to learn a predictive mapping from the textual domain to the corresponding taxonomic label space:
\begin{equation}
    \label{eq:representation}
    \vf_{tax}: \mathcal{T} \to \mathcal{Y}_{\text{smooth}}.
\end{equation}
Namely, for any input news text $T_i \in \mathcal{T}$, the model yields a prediction vector $\hat{\vy}_{tax}^{(i)}$ whose dimensionality aligns with that of $\mathcal{Y}_{\text{smooth}}$ (i.e., a 17-dimensional vector representing the combined taxonomic facets).

\paragraph{Adaptive gated pooling.}
For each input text $T_i$, we obtain token-level hidden states from a pre-trained LLM:
\begin{equation}
    \label{eq:llm_states}
    \mathbf{S}_i = \text{LLM}(T_i) \in \mathbb{R}^{L_i \times D},
\end{equation}
where $L_i$ is the sequence length and $D$ is the hidden dimensionality. Instead of relying on  mean or first-token pooling, we introduce a learnable gating vector that adaptively assigns importance weights to individual tokens:
\begin{equation}
    \label{eq:gate_pooling_res}
    \mathbf{v}_i = \sum_{h=1}^{L_i} g_{i,h}\mathbf{S}_{i,h} \in \mathbb{R}^{D},
\end{equation}
where $\mathbf{g}_i=\mathrm{softmax}(\mathbf{w}_{1:L_i})$ denotes the normalized token-level gating weights. This mechanism allows the model to learn a weighted aggregation of token features before taxonomic projection.

\paragraph{Taxonomy-aligned representation learning training.}
Given the pooled representation $\mathbf{v}_i \in \mathbb{R}^{D}$, we employ three facet-specific MLP heads, corresponding to Persuasion, Emotion, and Narrative Roles, to project the text into their respective taxonomic spaces:
\begin{equation}
\label{eq:taxonomy_heads}
\hat{\mathbf{p}}_i=f_P(\mathbf{v}_i),\quad
\hat{\mathbf{e}}_i=f_E(\mathbf{v}_i),\quad
\hat{\mathbf{n}}_i=f_N(\mathbf{v}_i),
\end{equation}
where each head outputs sigmoid-normalized probabilities over its corresponding taxonomy categories. The model is trained using Binary Cross-Entropy (BCE) supervision against the smoothed taxonomic targets:
\begin{equation}
\label{eq:taxonomy_loss}
\mathcal{L}_{\text{total}}
=
\mathcal{L}_P+\mathcal{L}_E+\mathcal{L}_N.
\end{equation}
Each facet-specific loss is computed independently over its corresponding category space using the smoothed labels derived from $\mathcal{Y}_{\text{smooth}}$.

\subsection{Taxonomy-based Disinformation Detection}
In the Taxonomy-based Disinformation Detection phase, we integrate the taxonomic feature vector $\mathbf{t}_i = \left[ \hat{\mathbf{p}}_i \; \| \; \hat{\mathbf{e}}_i \; \| \; \hat{\mathbf{n}}_i \right]$, formed by concatenating the attributes elicited from the taxonomic alignment tasks, with the text feature sequence $\mathbf{S}_i$. This fusion mechanism enables the model to leverage structured rhetorical, emotional, and narrative evidence when predicting whether a given news article constitutes disinformation.

\paragraph{Text feature transformation.}
To align contextual token features with the taxonomic space, we project each token representation into the same 17-dimensional space as the taxonomic attributes:
\begin{equation}
\label{eq:text_transformation}
    \tilde{\mathbf{S}}_{i,h}
    =
    \mathrm{TransMLP}(\mathbf{S}_{i,h})
    \in \mathbb{R}^{17},
    \quad h=1,\dots,L_i .
\end{equation}
The resulting sequence $\tilde{\mathbf{S}}_i \in \mathbb{R}^{L_i \times 17}$ serves as the contextual feature input for heterogeneous multi-head attention.
\begin{table*}[t] 
\centering
\small
\begin{tabular}{l cc cc cc}
\toprule
\multirow{2}{*}{\textbf{Method}} & \multicolumn{2}{c}{\textbf{Overall}} & \multicolumn{2}{c}{\textbf{Post}} & \multicolumn{2}{c}{\textbf{Article}} \\
\cmidrule(lr){2-3} \cmidrule(lr){4-5} \cmidrule(lr){6-7}
 & \textbf{MacroF1} & \textbf{MacroRecall} & \textbf{MacroF1} & \textbf{MacroRecall} & \textbf{MacroF1} & \textbf{MacroRecall} \\
\midrule
RAEmoLLM & $\underline{0.8255}_{\pm .002}$ & $0.7981_{\pm .002}$ & $0.8625_{\pm .001}$ & $0.8528_{\pm .001}$ & $0.7978_{\pm .002}$ & $0.7704_{\pm .002}$ \\
FACTUAL & $0.7563_{\pm .004}$ & $0.7377_{\pm .004}$ & $0.6428_{\pm .006}$ & $0.6474_{\pm .005}$ & $\underline{0.8027}_{\pm .004}$ & $0.7780_{\pm .004}$ \\
DYNAMO & $0.6680_{\pm .007}$ & $0.6972_{\pm .009}$ & $0.5737_{\pm .020}$ & $0.6200_{\pm .019}$ & $0.6936_{\pm .002}$ & $0.7177_{\pm .015}$ \\
PCoT & $0.8218_{\pm .020}$ & $0.8053_{\pm .019}$ & $0.7675_{\pm .007}$ & $0.7606_{\pm .007}$ & $\mathbf{0.8439}_{\pm .027}$ & $\mathbf{0.8226}_{\pm .025}$ \\
\midrule
RoBERTa-large & $0.8143_{\pm .028}$ & $\underline{0.8085}_{\pm .038}$ & $0.9270_{\pm .016}$ & $0.9300_{\pm .014}$ & $0.7583_{\pm .042}$ & $0.7517_{\pm .053}$ \\
DRES & $0.7294_{\pm .001}$ & $0.7170_{\pm .001}$ & $\underline{0.9454}_{\pm .002}$ & $\underline{0.9418}_{\pm .002}$ & $0.6194_{\pm .002}$ & $0.6145_{\pm .002}$ \\
MSynFD & $0.7337_{\pm .026}$ & $0.7206_{\pm .026}$ & $0.8887_{\pm .006}$ & $0.8924_{\pm .003}$ & $0.6375_{\pm .051}$ & $0.6356_{\pm .039}$ \\
\midrule
\textbf{ExTax (ours)} & $\mathbf{0.8456}_{\pm .011}$ & $\mathbf{0.8423}_{\pm .015}$ & $\mathbf{0.9548}_{\pm .003}$ & $\mathbf{0.9565}_{\pm .004}$ & $0.7965_{\pm .018}$ & $\underline{0.7913}_{\pm .022}$ \\
\bottomrule
\end{tabular}
\caption{Performance comparison between baseline models and our framework. The methods above the horizontal dividing line represent LLM-based approaches, while those below are deep learning-based methods. The best and second-best results are highlighted in \textbf{bold} and \underline{underlined}, respectively. We report the mean and standard deviation of the Macro $F_1$-score and Macro recall over all random seeds.}
\label{tab:main_results}
\end{table*}
\paragraph{Learnable taxonomic prompt.} To enrich the semantic representations of the taxonomic dimensions, we introduce a Learnable Taxonomic Prompt mechanism. Specifically, the previously obtained taxonomic feature vector $\mathbf{t}_i \in \mathbb{R}^{17}$ is concatenated with $N_{\text{ppt}}$ learnable semantic prompt vectors $\mathbf{q} \in \mathbb{R}^{N_{ppt} \times 17}$ of the same dimensionality, yielding an augmented representation:
\begin{equation}
    \tilde{\mathbf{t}}_i = \text{Concat} (\mathbf{t}_i, \mathbf{q}) \in  \mathbb{R}^{(1+N_{ppt}) \times 17}.
\end{equation}

This concatenation operation enables the model to adaptively recalibrate feature importance based on both the original taxonomic attributes and the learnable prompts. Consequently, the framework can more effectively capture nuanced semantic cues.

\paragraph{Heterogeneous multi-head attention.}
To maintain a strict one-to-one correspondence with the predetermined taxonomic spaces, we propose a Heterogeneous Multi-Head Attention mechanism. Unlike standard multi-head attention, where all heads share the same dimensionality, we configure three specialized attention heads whose dimensionalities correspond directly to the number of categories in the three taxonomic facets: $d_1 = |\mathcal{C}_P| = 6$, $d_2 = |\mathcal{C}_E| = 5$, and $d_3 = |\mathcal{C}_N| = 6$.

Formally, given the learnable taxonomic prompt sequence $\tilde{\mathbf{t}}_i$
and the transformed token feature sequence $\tilde{\mathbf{S}}_i$, we project
them into the query, key, and value spaces of the $k$-th taxonomic head:
$\mathbf{Q}_k \in \mathbb{R}^{(1+N_{ppt}) \times d_k}$,
$\mathbf{K}_k \in \mathbb{R}^{L_i \times d_k}$, and
$\mathbf{V}_k \in \mathbb{R}^{L_i \times d_k}$, where
$d_k \in \{|\mathcal{C}_P|, |\mathcal{C}_E|, |\mathcal{C}_N|\}$.
The heterogeneous attention output is computed as:
\begin{equation}
\label{eq:hetero_attn}
\begin{aligned}
\mathbf{H}_{i,k}
&= \mathrm{softmax}\left(
\frac{\mathbf{Q}_k \mathbf{K}_k^\top}{\sqrt{d_k}}
\right)\mathbf{V}_k, \\
\mathbf{H}_i^{\mathrm{het}}
&= \mathrm{Concat}\left(
\mathbf{H}_{i,1}, \mathbf{H}_{i,2}, \mathbf{H}_{i,3}
\right),
\end{aligned}
\end{equation}
where $\mathbf{H}_{i,k}$ denotes the output of the $k$-th attention head for instance $i$. The concatenated representation
$\mathbf{H}_i^{\mathrm{het}} \in \mathbb{R}^{(1+N_{ppt}) \times 17}$
combines the three facet-specific outputs along the feature dimension,
thereby preserving alignment with the unified 17-dimensional taxonomic space.

This design ensures that each attention head is dedicated to a single taxonomic dimension, preserving rigorous alignment with persuasion strategies, emotional manipulation, and narrative roles.

The heterogeneous multi-head attention module comprises a stack of $N_{\text{att}}$ layers. Within each layer, the input representation first undergoes heterogeneous multi-head attention, followed by a residual connection and layer normalization.

\paragraph{Prediction. } The classification head maps the final attention output $\mathbf{h}^{(N_{\text{att}})}_i \in \mathbb{R}^{(N_{\text{ppt}}+1) \times 17}$ into the target prediction space. Specifically, a Weighted Sum Queries layer first aggregates the representation across the query dimension to produce a unified 17-dimensional taxonomic vector, which is subsequently projected by a compact MLP layer to generate the final classification:
\label{eq:prediction_head}
\begin{gather}
    \mathbf{h}^{(sum)}_i = \sum_{j=1}^{N_{\text{ppt}}+1} \beta_j \cdot \mathbf{h}_{i,j}^{(N_{\text{att}})}, \\
    \hat{\vy_i} = \text{Linear}\big(\text{LN} (\text{GELU}(\text{Linear}(\mathbf{h}^{(sum)}_i))) \big),
\end{gather}
where $\beta_j$ denotes the learnable query weights, and $\hat{\mathbf{y}}$ represents the final prediction vector.

\section{Experiments}

\subsection{Setup}
\paragraph{Dataset.} We adopt the five datasets used in PCoT \citep{modzelewskiPCoTPersuasionAugmentedChain2025}: \textbf{CoAID} \citep{cuiCoAIDCOVID19Healthcare2020} and \textbf{ECTF} \citep{Paka2021CrossSEANAC, Bansal2021CombiningEA} cover COVID-19 health disinformation in articles and tweets, respectively, and provide train/validation/test splits; \textbf{ISOT Fake News} \citep{ahmedDetectingOpinionSpams2018, 10.1007/978-3-319-69155-8_9}, \textbf{MultiDis} \citep{modzelewskiPCoTPersuasionAugmentedChain2025}, and \textbf{EUDisinfo} \citep{modzelewskiPCoTPersuasionAugmentedChain2025} are test-only article corpora spanning mainstream news, multi-topic European political/social disinformation, and pro-Kremlin geopolitical narratives. Full specifications are provided in Appendix~\ref{app:datasets}.

\paragraph{Baseline.} We compare against seven representative baselines, including three deep learning-based methods, \textbf{RoBERTa} \citep{liuRoBERTaRobustlyOptimized2019}, \textbf{DRES} \citep{Farhangian2025DRESFN}, and \textbf{MSynFD} \citep{Xiao2024MSynFDMS}, and four LLM-based methods, \textbf{RAEmoLLM} \citep{liuRAEmoLLMRetrievalAugmented2025}, \textbf{FACTUAL} \citep{liMitigatingBiasesLarge2025}, \textbf{DYNAMO} \citep{jinDynamicKnowledgeUpdateDriven2025}, and \textbf{PCoT} \citep{modzelewskiPCoTPersuasionAugmentedChain2025}. Their introductions and implementation details are provided in Appendix~\ref{app:baseline-intro} and Appendix~\ref{app:baseline_imp}, respectively.

\paragraph{Metrics.} To quantitatively evaluate our framework, we employ two standard metrics: \textbf{Accuracy}, \textbf{Macro Recall} and \textbf{Macro $F_1$-score}. For a more detailed definition, please refer to Appendix~\ref{app:experiments}.
\begin{figure*}[t]
\centering
\includegraphics[width=\textwidth]{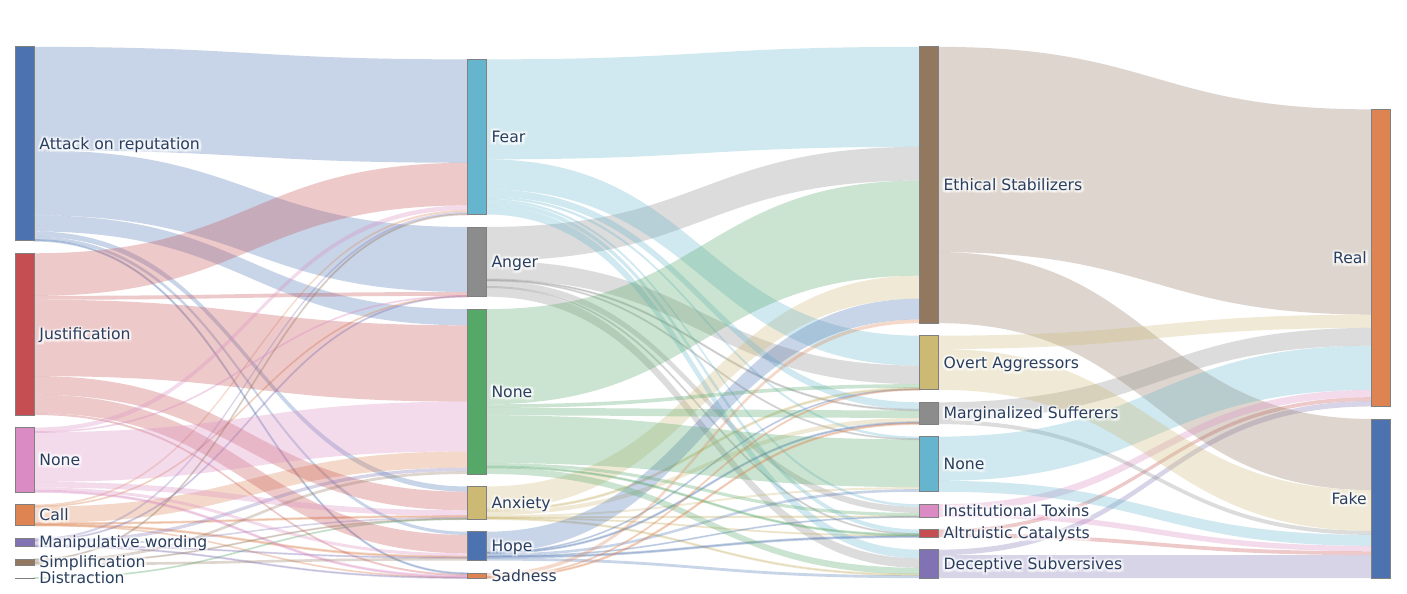}
\caption{Sankey diagram illustrating the co-occurrence patterns across the three taxonomic dimensions (Persuasion, Emotion, and Narrative Roles) and the final ground-truth labels. The flows reveal which manipulative attributes tend to appear together. "None" denotes instances where the LLMs did not detect any of the predefined categories within a specific taxonomic facet.}
\label{fig:sankey_diagram}
\end{figure*}
\paragraph{Implementation.} ExTax uses a frozen \texttt{roberta-large} backbone with parameter-efficient trainable components ($\approx$267K parameters), optimized in a two-phase architecture across three random seeds. All deep learning baselines and ExTax are trained and validated on the CoAID and ECTF training/validation partitions, then evaluated across all five test sets. Detailed implementation and hyperparameter configurations are provided in Appendix~\ref{app:extax_imp}.
\subsection{Main Results}
We categorize all five datasets into \textit{Post} and \textit{Article} genres based on text typology. Table~\ref{tab:main_results} reports the main results, from which we draw three key observations. 
(i) \textbf{Best Overall performance.} \textbf{ExTax} achieves the highest Overall MacroF1 (\textbf{0.8456}) and MacroRecall (\textbf{0.8423}), improving by $+2.01$ and $+4.42$ points over the strongest LLM-based RAG baseline (\texttt{RAEmoLLM}), and by $+3.13$ and $+3.38$ points over the strongest deep-learning baseline (\texttt{RoBERTa-large}). As noted by \citet{modzelewskiPCoTPersuasionAugmentedChain2025}, prompting-based baselines may further benefit from potential pre-training exposure, an advantage that ExTax does not require. 
(ii) \textbf{Genre trade-off rather than uniform dominance.} The Overall margin is mainly driven by short-form \textit{Posts}, where ExTax reaches \textbf{0.9548} MacroF1 and outperforms \texttt{PCoT} by $+18.7$ points. On long-form \textit{Articles}, ExTax is the runner-up with \textbf{0.7965} MacroF1, behind \texttt{PCoT} (\textbf{0.8439}), which may be attributable to the 30\% \textit{Article} minority in training and potential pre-training exposure favoring article-style text for prompting baselines. Crucially, \texttt{PCoT} pays for this \textit{Article} advantage by collapsing on \textit{Posts} (\textbf{0.7675}), whereas ExTax does not exhibit a genre-collapse failure mode. 
(iii) \textbf{Robustness under severe genre imbalance.} Despite the 70/30 genre imbalance, ExTax generalizes consistently across both genres, while conventional deep-learning baselines exhibit substantial performance degradation. For example, \texttt{DRES}'s MacroF1 drops from \textbf{0.9454} on \textit{Posts} to \textbf{0.6194} on \textit{Articles} ($-32.6$ points), suggesting that taxonomy-aligned features capture more genre-invariant manipulation patterns, whereas surface-pattern detectors may overfit to genre-specific artifacts.
\begin{table}[t]
\centering
\begin{tabular}{l c c} 
\toprule
\multirow{2}{*}{\textbf{LLM}} & \multicolumn{1}{c}{\textbf{stage 1}} & \textbf{stage 2} \\
\cmidrule(lr){2-2} \cmidrule(lr){3-3}
 & \textbf{Avg. F1} & \textbf{\begin{tabular}[c]{@{}c@{}}Overall\end{tabular}} \\
\midrule
GPT-5 mini          & $\mathbf{0.5026}$ & $0.8260$ \\
DeepSeek-V4-Flash     & $0.3234$ & $0.8216$ \\
Grok 4.1 Fast Reasoning         & $0.2571$ & $0.8384$ \\
Qwen3.6 Flash      & $0.4286$ & $\underline{0.8387}$ \\
\midrule
\textbf{ours (hybrid)} & $\underline{0.4489}$ & $\mathbf{0.8421}$ \\
\bottomrule
\end{tabular}
\caption{Performance comparison (Macro $F_1$) of our hybrid approach (Divergent Attribute Synthesis and Entropy-driven Dynamic Label Smoothing) against single LLM attribute synthesis across Stage 1 (Taxonomy-aligned Representation Learning) and Stage 2 (Taxonomy-based Disinformation Detection). Stage 1 reports the average Macro $F_1$ across the three taxonomic dimensions.}
\label{tab:syn_abla}
\end{table}
\begin{table}[t] 
\centering
\small
\begin{tabular}{lccc}
\toprule
\textbf{Conf.} & \textbf{$\mathbf{Acc}_{fake}$} & \textbf{$\mathbf{Acc}_{real}$} & \textbf{MacroF1} \\
\midrule
Baseline & $0.5952$ & $0.9313$ & $0.7789$ \\
Hetero. Attention & $0.6041$ & $0.9415$ & $0.7898$ \\
Ada. Gated Pooling & $0.6789$ & $0.8973$ & $0.7966$ \\
hybrid & $0.7132$ & $0.9001$ & $0.8137$ \\
$N_{\text{ppt}}=1$ & $0.5165$ & $\mathbf{0.9551}$ & $0.7541$ \\
$N_{\text{ppt}}=2$ & $\underline{0.7512}$ & $\underline{0.9517}$ & $\mathbf{0.8644}$ \\
$N_{\text{ppt}}=3$(Full) & $\mathbf{0.8058}$ & $0.8823$ & $\underline{0.8421}$ \\
\bottomrule
\end{tabular}
\caption{Ablation study of ExTax components. The baseline configuration utilizes RoBERTa paired with an MLP. Random seed is 445. Components are incrementally integrated from top to bottom until the full ExTax framework is achieved.}
\label{tab:ablation}
\end{table}
\begin{figure*}[t]
\centering
\includegraphics[width=\textwidth]{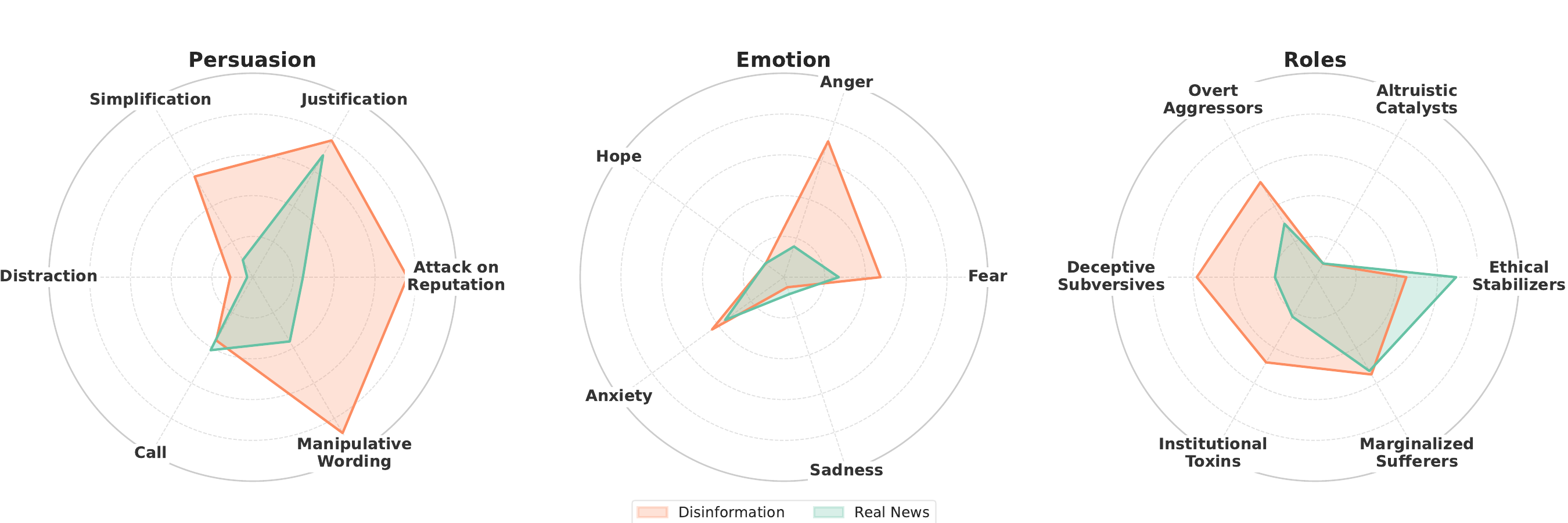}
\caption{Distribution of persuasion, emotion, and narrative role taxonomic attributes across all datasets, where distributions for disinformation and real news are highlighted in \textcolor{myred}{red} and \textcolor{mygreen}{green}, respectively.}
\label{fig:tax_distribution}
\end{figure*}
\subsection{Ablation}
As shown in Table~\ref{tab:syn_abla}, our hybrid Taxonomic Attribute Elicitation framework does not achieve the best Stage-1 performance in taxonomy-aligned representation learning, but yields the strongest downstream Macro $F_1$ in Stage 2. This is by design: Stage 1 favors hard-label fidelity to synthesized annotations, whereas entropy-driven dynamic label smoothing deliberately trades such fidelity for calibrated uncertainty. When multiple LLMs disagree, supervision is softened to prevent overfitting to uncertain attributes. As a result, Stage 2 receives better-calibrated taxonomic supervision, improving robustness against synthesized attribute noise and leading to stronger disinformation detection.

Table~\ref{tab:ablation} details the incremental contributions of ExTax's core components over the baseline (0.7789 macF1). Introducing heterogeneous multi-head attention effectively integrates taxonomic representations with contextual text features, improving performance to 0.7898 macF1. Adding adaptive gated pooling dynamically captures key semantic tokens, boosting fake news detection accuracy ($\text{Acc}_{\text{fake}}$) to 0.6789. Further integration of the hybrid module (divergent attribute synthesis and entropy-driven label smoothing) filters synthetic feature noise, increasing the macF1 score to 0.8137.

Regarding prompt scaling, although $N_{\text{ppt}}=2$ achieves the highest overall macF1 (0.8644), it suffers from severe classification bias toward the real label ($\text{Acc}_{\text{real}}=0.9517$ vs. $\text{Acc}_{\text{fake}}=0.7512$). Consequently, ExTax adopts $N_{\text{ppt}}=3$ to achieve more balanced diagnostic performance between fake ($\text{Acc}_{\text{fake}}=0.8058$) and real ($\text{Acc}_{\text{real}}=0.8823$) news, while maintaining a robust and competitive 0.8421 macF1 across varying seeds.

\subsection{Interpretability}
Figure~\ref{fig:tax_distribution} visualizes the global distribution of ExTax's three taxonomic dimensions across all datasets. Disinformation and real news exhibit clearly distinct patterns within the 17-dimensional taxonomic space: fake news is more strongly associated with ``Attack on Reputation'', ``Simplification'', and ``Manipulative Wording'', high-arousal emotions such as ``Anger'' and ``Fear'', and negative narrative roles such as ``Deceptive Subversives'' and ``Institutional Toxins''. These observations are consistent with prior findings on persuasion and emotional manipulation in disinformation \citep{modzelewskiPCoTPersuasionAugmentedChain2025,jamiesonFlaggingEmotionalManipulation2025,zhaoHowIndividualsCope2024}.

Figure~\ref{fig:sankey_diagram} further illustrates the co-occurrence structure among persuasion, emotion, narrative roles, and final labels. Aggressive persuasion frequently co-occurs with negative emotions, whereas non-manipulative attributes tend to propagate across dimensions and are more strongly associated with real news. These results suggest that ExTax captures interpretable manipulative patterns underlying deceptive texts. The full interpretability analysis and detailed case studies are provided in Appendix~\ref{app:interpretability}.

\section{Conclusion}
In this paper, we introduced \textbf{ExTax}, an explainable and taxonomy-aligned framework for disinformation detection that models three complementary cognitive facets of manipulation---persuasive rhetoric, emotional manipulation, and narrative roles---within a unified 17-dimensional taxonomic space. ExTax combines Divergent Attribute Elicitation, Entropy-driven Dynamic Label Smoothing, and Heterogeneous Multi-Head Attention to produce noise-robust and structurally aligned representations of manipulative intent. Across five cross-domain, cross-genre datasets, ExTax outperforms strong LLM- and deep-learning baselines in Overall Macro $F_1$ and remains robust under severe genre imbalance. More importantly, by exposing per-sample 17-dimensional manipulation profiles, ExTax moves disinformation detection beyond binary fake/real verdicts toward an interpretable and auditable diagnosis of how deceptive content manipulates readers, thereby providing a more transparent foundation for downstream fact-checking and content-moderation workflows.

\section*{Limitations}
Several further limitations are worth noting. (1) \textbf{Reliance on commercial frontier LLMs.} The divergent attribute elicitation stage relies on four commercial frontier LLMs, which incurs non-trivial inference cost and raises reproducibility concerns as those APIs evolve. Although entropy-driven smoothing mitigates inconsistency across models, it cannot correct systematic biases shared by all of them. (2) \textbf{English- and Western-centric corpora.} Our five evaluation datasets are predominantly English and focused on Western political and health discourse; thus, we do not validate ExTax on multilingual or non-Western disinformation, where persuasion conventions and narrative archetypes may differ. (3) \textbf{Explanations are not yet evaluated with humans.} While ExTax outputs structured per-sample 17-dimensional manipulation profiles, we have not conducted a user study with fact-checkers or moderators to verify whether these profiles are more useful in practice than opaque verdicts. We leave controlled human evaluation to future work.

Although ExTax is designed to support explainable disinformation detection, it may still produce false positives or false negatives, especially on long-form articles or domains underrepresented in the training data. False positives could unfairly flag legitimate content as disinformation, while false negatives could allow harmful misleading content to remain undetected. In addition, the LLM-elicited taxonomic attributes may reflect biases or inconsistencies from the underlying LLMs, even though our entropy-driven smoothing mechanism is designed to mitigate synthetic labeling noise. Therefore, ExTax should be used as a decision-support tool rather than as an automated content moderation system, and its outputs should be interpreted with human oversight.


\bibliography{custom}

\newpage
\appendix

\section*{Appendix}

\section{Taxonomies}
\label{app:taxonomies}

This appendix provides the detailed definitions of the three taxonomic dimensions used in ExTax: Persuasion, Emotion, and Narrative Roles. These taxonomies are used during the Taxonomic Attribute Elicitation stage to guide LLMs in extracting interpretable textual attributes from each input text. The definitions are derived from the prompt configuration used for attribute elicitation, while prompt-formatting instructions and JSON output templates are reported separately in Appendix~\ref{app:prompts}.

\subsection{Persuasion}

Persuasion refers to the use of language intended to influence readers. We distinguish six high-level persuasion approaches: Attack on Reputation, Justification, Simplification, Distraction, Call, and Manipulative Wording.

\subsubsection{Attack on Reputation}
Attack on Reputation refers to arguments that do not directly address the topic itself, but instead target a participant, group, organization, object, or activity in order to question or undermine its credibility.
\begin{itemize}
    \item \textbf{Name Calling or Labelling:} The use of loaded labels directed at an individual, group, object, or activity, typically in an insulting or demeaning way, although sometimes using labels that the target audience may find desirable.
    \item \textbf{Guilt by Association:} Attacking an opponent, activity, or concept by associating it with another group, activity, or concept that carries strong negative connotations for the target audience.
    \item \textbf{Casting Doubt:} Questioning the character or personal attributes of a person or entity in order to undermine its general credibility or perceived quality.
    \item \textbf{Appeal to Hypocrisy:} Attacking a target's reputation by accusing it of hypocrisy or inconsistency.
    \item \textbf{Questioning the Reputation:} Making strong negative claims about a target, with emphasis on undermining its character or moral standing rather than addressing the substantive topic.
\end{itemize}

\subsubsection{Justification}
Justification refers to arguments composed of a statement and an accompanying explanation or appeal, where the latter is used to justify or support the former.
\begin{itemize}
    \item \textbf{Flag Waving:} Justifying an idea by invoking pride in a group or highlighting benefits for that group.
    \item \textbf{Appeal to Authority:} Giving weight to an argument, idea, or piece of information by presenting an entity regarded as authoritative as its source.
    \item \textbf{Appeal to Popularity:} Justifying an argument or idea by claiming that everyone, or the large majority, agrees with it, or that nobody disagrees with it.
    \item \textbf{Appeal to Values:} Supporting an idea by linking it to values that are viewed positively by the target audience.
    \item \textbf{Appeal to Fear or Prejudice:} Promoting or rejecting an idea by invoking fear, prejudice, or repulsion toward that idea.
\end{itemize}

\subsubsection{Simplification}
Simplification refers to arguments that excessively simplify a complex issue, usually by reducing its causes, consequences, or available choices.
\begin{itemize}
    \item \textbf{Causal Oversimplification:} Assuming a single cause or reason for an issue when multiple causes are involved.
    \item \textbf{False Dilemma or No Choice:} Presenting only two options or sides when more alternatives exist, or explicitly instructing the audience to take one action while excluding other possible choices.
    \item \textbf{Consequential Oversimplification:} Claiming that an initial event or action will lead to a domino-like chain of increasingly serious consequences, where the causal chain is exaggerated, improbable, or insufficiently warranted.
\end{itemize}

\subsubsection{Distraction}
Distraction refers to arguments that divert attention away from the main issue or argument.
\begin{itemize}
    \item \textbf{Strawman:} Giving the impression of refuting an opponent's argument while replacing the actual argument with a distorted or weaker version.
    \item \textbf{Red Herring:} Diverting the audience's attention from the main topic by introducing an irrelevant issue.
    \item \textbf{Whataboutism:} Attempting to discredit an opponent's position by accusing the opponent of hypocrisy without directly addressing or disproving the original argument.
\end{itemize}

\subsubsection{Call}
Call refers to language that is not primarily argumentative, but instead encourages the audience to act or think in a particular way.
\begin{itemize}
    \item \textbf{Slogans:} Brief and striking phrases, often functioning as emotional appeals and potentially involving labelling or stereotyping.
    \item \textbf{Conversation Killer:} Words or phrases that discourage critical thinking or meaningful discussion about a given topic.
    \item \textbf{Appeal to Time:} Arguments centered on the claim that the time has come for a particular action.
\end{itemize}

\subsubsection{Manipulative Wording}
Manipulative Wording refers to language that is not necessarily an argument by itself, but uses non-neutral, confusing, exaggerated, or loaded expressions to influence the reader emotionally.
\begin{itemize}
    \item \textbf{Loaded Language:} The use of words or phrases with strong emotional implications, either positive or negative, to influence the audience.
    \item \textbf{Obfuscation, Intentional Vagueness, or Confusion:} The use of deliberately unclear, vague, or ambiguous language that leaves room for audience-specific interpretations.
    \item \textbf{Exaggeration or Minimisation:} Representing something in an excessive manner or making it appear less important or smaller than it actually is.
    \item \textbf{Repetition:} Repeated use of the same phrase or expression in order to increase persuasive impact.
\end{itemize}

\subsection{Emotion}

Emotion refers to language designed to bypass purely rational processing and influence readers by triggering affective responses. We distinguish five high-level emotional manipulation types: Fear, Anger, Hope, Anxiety, and Sadness.

\begin{itemize}
    \item \textbf{Fear:} Language designed to make readers feel scared or threatened, often by emphasizing immediate risks, serious health concerns, or distressing outcomes in order to provoke a protective reaction.
    \item \textbf{Anger:} Language designed to provoke outrage, resentment, or a sense of injustice, often by targeting specific groups, institutions, or actors.
    \item \textbf{Hope:} Language designed to elicit positive expectations or promise breakthroughs, often by presenting unverified cures, revolutionary alternatives, or highly optimistic outcomes.
    \item \textbf{Anxiety:} Language designed to create uncertainty about future safety or well-being, often by describing disturbing cases or hidden dangers in everyday practices.
    \item \textbf{Sadness:} Language focusing on irreversible loss, heartbreaking misfortune, or the suffering of helpless victims, thereby triggering sympathy, sorrow, or moral guilt.
\end{itemize}

\subsection{Narrative Roles}

Narrative Roles describe how a person, group, or entity is functionally positioned within a narrative, independently of its actual moral status. We distinguish six high-level role categories: Ethical Stabilizers, Altruistic Catalysts, Overt Aggressors, Deceptive Subversives, Institutional Toxins, and Marginalized Sufferers.

\subsubsection{Ethical Stabilizers}
Ethical Stabilizers are actors portrayed as using established values, moral authority, or mediation to maintain social cohesion, safety, and justice.
\begin{itemize}
    \item \textbf{Guardian:} Actors portrayed as protectors of values or communities, such as law enforcement officers, soldiers, or community leaders.
    \item \textbf{Peacemaker:} Actors portrayed as seeking harmony, conflict resolution, diplomacy, negotiation, or mediation.
    \item \textbf{Virtuous:} Actors portrayed as righteous, noble, fair, just, or morally exemplary.
\end{itemize}

\subsubsection{Altruistic Catalysts}
Altruistic Catalysts are actors portrayed as challenging the status quo or enduring hardship to drive systemic change, liberation, or collective benefit.
\begin{itemize}
    \item \textbf{Rebel:} Actors portrayed as challenging the status quo and fighting for major change, freedom, or liberation from oppression.
    \item \textbf{Martyr:} Actors portrayed as sacrificing their well-being, or even their lives, for a greater good or cause.
    \item \textbf{Underdog:} Actors portrayed as disadvantaged or unlikely to succeed, but striving against stronger forces or obstacles.
\end{itemize}

\subsubsection{Overt Aggressors}
Overt Aggressors are visible antagonists portrayed as using direct force, systemic power, or violent ideology to dominate, divide, or harm others.
\begin{itemize}
    \item \textbf{Instigator:} Actors portrayed as initiating conflict, tension, unrest, or violence.
    \item \textbf{Tyrant:} Actors portrayed as abusing power, ruling unjustly, or oppressing those under their control.
    \item \textbf{Terrorist:} Actors portrayed as engaging in violence or terror to advance ideological goals, often by targeting civilians.
    \item \textbf{Bigot:} Actors portrayed as hostile or discriminatory toward specific groups, including racism, sexism, homophobia, antisemitism, Islamophobia, or other forms of hate speech.
\end{itemize}

\subsubsection{Deceptive Subversives}
Deceptive Subversives are hidden or manipulative actors portrayed as undermining trust, truth, or safety through secrecy, betrayal, deception, or strategic manipulation.
\begin{itemize}
    \item \textbf{Conspirator:} Actors portrayed as participating in secret plots or covert plans to deceive or undermine others.
    \item \textbf{Spy:} Actors portrayed as gathering or transmitting sensitive information to a rival or enemy through secrecy or deception.
    \item \textbf{Traitor:} Actors portrayed as betraying a cause, community, or country.
    \item \textbf{Deceiver:} Actors portrayed as manipulating truth, spreading misinformation, or shaping public perception for strategic or self-interested purposes.
    \item \textbf{Saboteur:} Actors portrayed as deliberately damaging, obstructing, or weakening systems, organizations, or processes.
    \item \textbf{Foreign Adversary:} External actors portrayed as creating geopolitical tension or acting against another country's interests.
\end{itemize}

\subsubsection{Institutional Toxins}
Institutional Toxins are actors or entities portrayed as internal sources of organizational or systemic failure, caused by vice, incapacity, corruption, or incompetence.
\begin{itemize}
    \item \textbf{Corrupt:} Actors portrayed as engaging in unethical or illegal behavior for personal, political, or financial gain.
    \item \textbf{Incompetent:} Actors portrayed as causing harm through ignorance, poor judgment, lack of skill, or inability to perform their duties.
\end{itemize}

\subsubsection{Marginalized Sufferers}
Marginalized Sufferers are actors portrayed as enduring harm, exploitation, neglect, exclusion, or unjust blame, often functioning as evidence of systemic failure or as targets for rescue.
\begin{itemize}
    \item \textbf{Forgotten:} Marginalized or overlooked groups portrayed as ignored by society and lacking needed attention or support.
    \item \textbf{Exploited:} Individuals or groups portrayed as being used for others' benefit without adequate consent and with harm to their well-being.
    \item \textbf{Victim:} Actors portrayed as suffering physical or economic harm due to circumstances beyond their control.
    \item \textbf{Scapegoat:} Actors portrayed as being blamed unjustly for problems or failures in order to divert attention from the real causes or responsible parties.
\end{itemize}

\section{Prompts}
\label{app:prompts}

The following section delineates the construction of prompts designed to guide LLMs in extracting taxonomic features across Persuasion, Emotion, and Narrative Roles. Specifically, the prompt configuration for eliciting Persuasion attributes is adopted from \citet{modzelewskiPCoTPersuasionAugmentedChain2025}. Uppercase placeholders wrapped in hashtags, such as \#PERSUASION\_DEFINITIONS\#, \#EMOTION\_DEFINITIONS\#, and \#NARRATIVE\_ROLE\_DEFINITIONS\#, are used as designators to denote the comprehensive taxonomic definitions detailed in Appendix~\ref{app:taxonomies}.

\subsection{Prompt 1: Extract Persuasion Taxonomic Features}

\begin{systemprompt}
You are an assistant who detects persuasion in text. Persuasive text is characterized by a specific use of language in order to influence readers. We distinguish the following high-level persuasion approaches:
    #PERSUASION_DEFINITIONS#

You are the expert who detects high-level persuasion approaches: Attack on Reputation, Justification, Simplification, Distraction, Call, Manipulative Wording.
\end{systemprompt}

\begin{contextprompt}
Analyze the text and decide if the text contains any high-level persuasion approaches from the following: Attack on Reputation, Justification, Simplification, Distraction, Call, Manipulative Wording. For each high-level persuasion approach, provide an explanation of how it appears in the analyzed text. Be conservative in your final decisions; when you are not fully sure, answer No. Give your answer in the form of a dictionary:
{{
"Attack_on_reputation": {{"is_used": "Your answer. Use only Yes or No", "explanation": "If high-level persuasion Attack on Reputation appears, provide an explanation here."}},
"Justification": {{"is_used": "Your answer. Use only Yes or No", "explanation": "If high-level persuasion Justification appears, provide an explanation here."}},
"Simplification": {{"is_used": "Your answer. Use only Yes or No", "explanation": "If high-level persuasion Simplification appears, provide an explanation here."}},
"Distraction": {{"is_used": "Your answer. Use only Yes or No", "explanation": "If high-level persuasion Distraction appears, provide an explanation here."}},
"Call": {{"is_used": "Your answer. Use only Yes or No", "explanation": "If high-level persuasion Call appears, provide an explanation here."}},
"Manipulative_wording": {{"is_used": "Your answer. Use only Yes or No", "explanation": "If high-level persuasion Manipulative Wording appears, provide an explanation here."}}
}}
Text: {text}
Answer:
\end{contextprompt}

\subsection{Prompt 2: Extract Emotion Taxonomic Features}

\begin{systemprompt}
You are a specialized linguistic analyst and expert in detecting emotional manipulation in online text. Your goal is to identify techniques designed to bypass rational thinking and influence readers by triggering specific emotional responses.

We distinguish between the following five high-level emotional appeals:
    #EMOTION_DEFINITIONS#

You are the expert who detects emotional manipulation types: Fear, Anger, Hope, Anxiety, Sadness.
\end{systemprompt}

\begin{contextprompt}
Analyze the text and decide if the text contains any emotional manipulation types from the following: Fear, Anger, Hope, Anxiety, Sadness. For each emotional manipulation type, provide an explanation of how it appears in the analyzed text. Be conservative in your final decisions; when you are not fully sure, answer No.

Give your answer in the form of a dictionary:
{{
"Fear": {{"is_used": "Your answer. Use only Yes or No", "explanation": "If the Fear type appears, provide a one-sentence explanation starting with 'This post contains language that may be intended to make you feel fear by...'"}},
"Anger": {{"is_used": "Your answer. Use only Yes or No", "explanation": "If the Anger type appears, provide a one-sentence explanation starting with 'This post contains language that may be intended to make you feel anger by...'"}},
"Hope": {{"is_used": "Your answer. Use only Yes or No", "explanation": "If the Hope type appears, provide a one-sentence explanation starting with 'This post contains language that may be intended to make you feel hope by...'"}},
"Anxiety": {{"is_used": "Your answer. Use only Yes or No", "explanation": "If the Anxiety type appears, provide a one-sentence explanation starting with 'This post contains language that may be intended to make you feel anxiety by...'"}},
"Sadness": {{"is_used": "Your answer. Use only Yes or No", "explanation": "If the Sadness type appears, provide a one-sentence explanation starting with 'This post contains language that may be intended to make you feel sadness by...'"}}
}}
Text: {text}
Answer:
\end{contextprompt}

\subsection{Prompt 3: Extract Narrative Role Taxonomic Features}

\begin{systemprompt}
You are an expert at identifying narrative framing and role portrayal in text. Your goal is to detect whether a text utilizes specific high-level narrative categories -- Ethical Stabilizers, Altruistic Catalysts, Overt Aggressors, Deceptive Subversives, Institutional Toxins, or Marginalized Sufferers -- to frame the participants in a story.

This analysis focuses on functional roles -- how a person, group, or entity is positioned within the narrative -- independent of their actual moral standing.

You distinguish between six high-level main roles based on the following archetypal markers:
    #NARRATIVE_ROLE_DEFINITIONS#

Your analysis must be based on the specific linguistic cues and narrative structure within the text.
\end{systemprompt}

\begin{contextprompt}
Analyze the text below to determine whether it contains any of the following high-level roles: Ethical Stabilizers, Altruistic Catalysts, Overt Aggressors, Deceptive Subversives, Institutional Toxins, or Marginalized Sufferers.

For each detected role, provide a concise explanation of how it is portrayed based on the narrative function and word choice. Be conservative in your final decisions; if you are not fully sure, answer No.

Return your answer as a JSON object in the following format:
{{
  "Ethical_Stabilizers": {{
    "is_used": "Yes or No",
    "explanation": "If Yes, provide evidence of how a participant is framed as a protector of values or a righteous figure (e.g., Guardian, Virtuous, Peacemaker)."
  }},
  "Altruistic_Catalysts": {{
    "is_used": "Yes or No",
    "explanation": "If Yes, provide evidence of how a participant is framed as driving positive change or sacrifice (e.g., Rebel, Martyr, Underdog)."
  }},
  "Overt_Aggressors": {{
    "is_used": "Yes or No",
    "explanation": "If Yes, provide evidence of how a participant is framed as an initiator of conflict or violence (e.g., Tyrant, Terrorist, Instigator, Bigot)."
  }},
  "Deceptive_Subversives": {{
    "is_used": "Yes or No",
    "explanation": "If Yes, provide evidence of how a participant is framed through secrecy, betrayal, or misinformation (e.g., Conspirator, Spy, Traitor, Deceiver, Saboteur, Foreign Adversary)."
  }},
  "Institutional_Toxins": {{
    "is_used": "Yes or No",
    "explanation": "If Yes, provide evidence of how a participant is framed as abusing power or as a source of systemic failure (e.g., Corrupt, Incompetent)."
  }},
  "Marginalized_Sufferers": {{
    "is_used": "Yes or No",
    "explanation": "If Yes, provide evidence of how a participant is framed as suffering harm, exploitation, neglect, exclusion, or unjust blame (e.g., Victim, Exploited, Forgotten, Scapegoat)."
  }}
}}

Text: {text}
Answer:
\end{contextprompt}

\section{Method}
\label{app:method}
\paragraph{Details of entropy-driven dynamic label smoothing.}
For each taxonomic facet, the binary LLM outputs are first aggregated by category. In the Emotion task, for example, $e_{i,k}^{(c)} \in \{0,1\}$ indicates whether the $k$-th LLM assigns sample $T_i$ to emotion category $c \in \mathcal{C}_E$. The positive vote proportion $p_{i,c}$ measures the fraction of LLMs predicting the presence of category $c$. We compute binary entropy over this Bernoulli distribution to quantify model disagreement. The same computation is applied to the Persuasion and Narrative Role facets by replacing $e_{i,k}^{(c)}$ with the corresponding components of $\mathbf{p}_{i,k}$ and $\mathbf{n}_{i,k}$. The resulting smoothed labels across all samples and categories form:
\begin{equation}
\label{eq:fine_tax_target}
\begin{split}
    \mathcal{Y}_{\text{smooth}} = \Big\{ y_{\text{tax}}^{(i,c)} \;\Big|\; & i=1,\dots,n, \\
    & c \in \mathcal{C}_P \cup \mathcal{C}_E \cup \mathcal{C}_N \Big\}.
\end{split}
\end{equation}
When all LLMs agree, $H_{i,c}=0$ and the smoothed label equals the vote proportion $p_{i,c}$. When the models disagree maximally, $H_{i,c}$ is high and the target is shifted toward $0.5$, reducing the effect of uncertain synthetic supervision.

\paragraph{Details of adaptive gated pooling.}
Given the LLM hidden states $\mathbf{S}_i \in \mathbb{R}^{L_i \times D}$, Adaptive Gated Pooling uses a learnable parameter vector $\mathbf{w} \in \mathbb{R}^{L_{\max}}$, where $L_{\max}$ is the maximum truncation length. For an input with length $L_i$, we slice the first $L_i$ elements of $\mathbf{w}$ as $\mathbf{w}_{1:L_i}$ and compute the normalized gating weights:
\begin{align}
\mathbf{g}_i &= \mathrm{softmax}(\mathbf{w}_{1:L_i}) \in \mathbb{R}^{L_i},
    \\
    g_{i,h} &= \frac{\exp(w_h)}{\sum_{j=1}^{L_i}\exp(w_j)}.    
\end{align}
The pooled representation is then obtained by a weighted sum:
\begin{equation}
    \mathbf{v}_i = \sum_{h=1}^{L_i} g_{i,h}\mathbf{S}_{i,h}.
\end{equation}
All elements of $\mathbf{w}$ are initialized to 1, making the initial pooling behavior equivalent to average pooling. During training, the gating vector is optimized to assign higher weights to more informative token positions.

\paragraph{Details of taxonomy-aligned representation learning.}
Each taxonomic facet is modeled using an independent MLP head:
\begin{equation}
\label{eq:mlp_structure}
\begin{split}
\mathbf{h}_i^{(c)}
&=
\mathrm{LN}
\big(
\mathrm{GELU}
(
\mathrm{Linear}_1^{(c)}(\mathbf v_i)
)
\big), \\
f_c(\mathbf v_i)
&=
\mathrm{Linear}_2^{(c)}
\big(
\mathrm{Dropout}(\mathbf h_i^{(c)})
\big),
\end{split}
\end{equation}
where
$\mathrm{Linear}_1^{(c)}:\mathbb R^D\to\mathbb R^{D_h}$
and
$\mathrm{Linear}_2^{(c)}:\mathbb R^{D_h}\to\mathbb R^{|\mathcal C_c|}$.
The outputs are passed through sigmoid activations to obtain multi-label probabilities.

For each facet, we optimize Binary Cross-Entropy (BCE) loss against the smoothed taxonomic labels. Taking Persuasion as an example:
\begin{equation}
\label{eq:lp_split}
\begin{split}
\mathcal L_P
=
-\frac1n
\sum_{i=1}^n
\sum_{c\in\mathcal C_P}
\Big[
&
y_{\text{tax}}^{(i,c)}
\log(\hat p_{i,c})
\\
+
&
(1-y_{\text{tax}}^{(i,c)})
\log(1-\hat p_{i,c})
\Big].
\end{split}
\end{equation}
The Emotion and Narrative Role losses are defined analogously.

\paragraph{Details of text feature transformation.}
The transformation module maps each token vector $\mathbf{S}_{i,h} \in \mathbb{R}^D$ into the 17-dimensional taxonomic space using a two-layer MLP:
\begin{equation}
\label{eq:trans_mlp_detail}
\begin{split}
\mathbf{h}_{\mathrm{mid}}
&=
\mathrm{Linear}_1
\big(
\mathrm{LN}(\mathbf{S}_{i,h})
\big), \\
\tilde{\mathbf{S}}_{i,h}
&=
\mathrm{Linear}_2
\big(
\mathrm{GELU}(\mathbf{h}_{\mathrm{mid}})
\big),
\end{split}
\end{equation}
where $\mathrm{Linear}_1:\mathbb{R}^D\to\mathbb{R}^{d_{\mathrm{ff}}}$ and $\mathrm{Linear}_2:\mathbb{R}^{d_{\mathrm{ff}}}\to\mathbb{R}^{17}$.

\section{Experiments}
\label{app:experiments}
\subsection{Dataset}
\label{app:datasets}
\begin{itemize}
\item \textbf{CoAID} \citep{cuiCoAIDCOVID19Healthcare2020}: A COVID-19 health disinformation dataset containing both news articles and social media posts, including over 4,000 news articles and more than 1,000 posts mainly from Twitter. The content covers pandemic-related topics such as public health policies, vaccines, treatment measures, and other healthcare claims. In our setting, it contains 5,011 samples across train, validation, and test splits, with a relatively stable genre distribution in each split: posts account for approximately 44\%, while articles account for approximately 56\%.

\item \textbf{ISOT Fake News} \citep{ahmedDetectingOpinionSpams2018, 10.1007/978-3-319-69155-8_9}: A large-scale fake news article dataset containing over 44,000 news articles collected from both reliable mainstream media, such as BBC and CNN, and unreliable or fake news websites. The articles span multiple domains, including politics, entertainment, technology, and health. Following the setting in PCoT, we use its test split with 499 samples, including 274 fake news instances and 225 real news instances.

\item \textbf{ECTF} \citep{Paka2021CrossSEANAC, Bansal2021CombiningEA}: An extended version of the CTF (COVID-19 Tweet Fake) dataset for COVID-19 fake tweet detection. It consists of tweets annotated as fake or real and covers pandemic-related topics such as virus transmission, protective measures, public responses, and vaccine controversies. This dataset provides train, validation, and test splits. In our setting, it contains 4,152 samples, with 3,001, 751, and 400 samples in the train, validation, and test sets, respectively.

\item \textbf{MultiDis} \citep{modzelewskiPCoTPersuasionAugmentedChain2025}: A researcher-constructed multi-topic disinformation dataset containing nearly 2,000 English news articles. The articles cover a broad range of political and social topics, including Anti-Europeanism, Anti-Atlanticism, Anti-migration, Xenophobia, Climate Change, Energy Crisis, Health, Institutional and Media Distrust, Gender Issues, Ukraine War, Refugees, and LGBT+. In our setting, this dataset is used as a test-only dataset and contains 499 test samples.

\item \textbf{EUDisinfo} \citep{modzelewskiPCoTPersuasionAugmentedChain2025}: A disinformation dataset built upon the EUvsDisinfo database, focusing on pro-Kremlin propaganda narratives and political disinformation. The articles mainly address issues such as EU policies, the situation in Ukraine, NATO expansion, and related geopolitical narratives. In our setting, this dataset is used as a test-only dataset and contains 359 articles, including 241 credible articles and 118 disinformation articles.
\end{itemize}

\subsection{Baseline Introduction}
\label{app:baseline-intro}
\begin{itemize}
\item \textbf{RoBERTa} \citep{liuRoBERTaRobustlyOptimized2019}: A BERT variant optimized via dynamic masking, larger batches, and expanded corpora without Next Sentence Prediction, serving as a competitive contextualized baseline.

\item \textbf{DRES} \citep{Farhangian2025DRESFN}: A framework routing instances to optimal feature spaces based on hardness measures to dynamically ensemble classifiers.

\item \textbf{MSynFD} \citep{Xiao2024MSynFDMS}: A syntax-aware framework capturing long-range dependency graphs and debiasing topical priors to detect fine-grained contextual distortions.

\item \textbf{RAEmoLLM} \citep{liuRAEmoLLMRetrievalAugmented2025}: An emotion-based retrieval-augmented framework using a specialized LLM to dynamically select affective exemplars for cross-domain detection.

\item \textbf{FACTUAL} \citep{liMitigatingBiasesLarge2025}: A counterfactual augmented calibration framework disentangling robust features from domain shortcuts to dynamically calibrate model confidence scores.

\item \textbf{DYNAMO} \citep{jinDynamicKnowledgeUpdateDriven2025}: A framework utilizing Monte Carlo Tree Search to verify factual paths and dynamically update knowledge graphs for real-time news detection.

\item \textbf{PCoT} \citep{modzelewskiPCoTPersuasionAugmentedChain2025}: A zero-shot prompting method that first analyzes persuasive strategies in the input text and then uses this persuasion-aware reasoning to detect disinformation.
\end{itemize}

\subsection{Metrics}
To quantitatively evaluate our framework, we employ two standard metrics: \textbf{Accuracy} and \textbf{Macro $F_1$-score}.

\textbf{Accuracy} measures the global proportion of correctly classified instances:
\begin{equation}
\text{Accuracy} = \frac{\text{TP} + \text{TN}}{\text{TP} + \text{TN} + \text{FP} + \text{FN}}
\end{equation}
where $\text{TP}$, $\text{TN}$, $\text{FP}$, and $\text{FN}$ represent True Positives, True Negatives, False Positives, and False Negatives, respectively.

To account for potential class imbalances, we also use the \textbf{Macro $F_1$-score}. It calculates the $F_1$-score for each class independently and takes their unweighted arithmetic mean. For a specific class $k$, its precision ($P_k$), recall ($R_k$), and $F_1$-score ($F_{1,k}$) are defined as:
\begin{align}
P_k &= \frac{\text{TP}_k}{\text{TP}_k + \text{FP}_k} \\
\quad R_k &= \frac{\text{TP}_k}{\text{TP}_k + \text{FN}_k} \\ \quad F_{1,k} &= \frac{2 \cdot P_k \cdot R_k}{P_k + R_k}
\end{align}
Given the binary class set $\mathcal{C} = \{0, 1\}$ (denoting reliable news and disinformation), the final Macro $F_1$-score is computed as:
\begin{equation}
\text{Macro } F_1 = \frac{1}{|\mathcal{C}|} \sum_{k \in \mathcal{C}} F_{1,k}
\end{equation}

Similarly, we report the \textbf{Macro Recall} to evaluate the model's ability to identify instances from each class regardless of class frequency. Macro Recall computes the recall for each class independently and then takes their unweighted arithmetic mean. Given the binary class set $\mathcal{C} = \{0, 1\}$, Macro Recall is defined as:

\begin{equation}
\text{Macro Recall} = \frac{1}{|\mathcal{C}|} \sum_{k \in \mathcal{C}} R_k
\end{equation}

\subsection{ExTax Implementation}
\label{app:extax_imp}
During the training process, we perform an exhaustive hyperparameter search to ensure optimal convergence. To guarantee that our empirical findings are statistically significant and robust, we conduct multiple independent training and evaluation runs across three distinct random seeds: 43, 434 and 445. We subsequently report the mean and standard deviation of the core performance metrics. For the divergent attribute synthesis operation, we leverage a diverse suite of frontier models, including GPT-5-mini, DeepSeek-V4-flash, Qwen3.6-flash, and Grok-4.1-fast-reasoning. The label smoothing hyperparameter $\alpha$ is strictly grounded at 0.1896 to mitigate potential decision noise. All experiments and model training phases are computationally lightweight and can be fully executed on a single NVIDIA RTX 4090 (24 GB VRAM) GPU. For further details regarding the comprehensive training configurations and architectural model profiles, please refer to Table~\ref{tab:model_settings} and Table~\ref{tab:train_settings}, respectively.

\begin{table*}[t] 
\centering
\small
\begin{tabular}{llp{7cm}}
\toprule
\textbf{Component} & \textbf{Parameter} & \textbf{Value} \\
\midrule
Taxonomies-Aligned & Adopted model & RoBERTa-large \citep{liuRoBERTaRobustlyOptimized2019} \\
           & Maximum token length & 512 \\
           & Pooling strategy & Adaptive gated pooling \\
           & Adaptive initialization & initialized to $1$ \\
           & Aligned MLP hidden dims & $[786, 32, 32]$ \\
           & Aligned MLP dropout rate & $0.3290$ \\
           & Aligned MLP output dimension & Persuasion\&Narrative Role: $d = 6$, Emotion: $d=5$ \\
\midrule
Disinformation Detection     & Number of HeteroAttention Block & $N_{att}=1$ (hidden dim = 6+6+5 = 17) \\
           & Classification Head & hidden dim = 128 \\
           & Activation & GELU \\
\midrule
\multicolumn{2}{l}{Total parameters} & $\approx$ 357\text{M}($\approx$ 267K trainable) \\
\bottomrule
\end{tabular}
\caption{Full architecture details of Extax}
\label{tab:model_settings}
\end{table*}

\begin{table*}[t] 
\centering
\small
\begin{tabular}{llp{7cm}}
\toprule
\textbf{Category} & \textbf{Parameter} & \textbf{Value} \\
\midrule
Taxonomies-Aligned & lr & $0.00065$ \\
           & Weight decay & $0.00069$ \\
           & Epochs & $10$ \\
           & Patience & $7$ \\
           & Seed & $[43, 434, 445]$ \\
           & Batch size & $128$ \\
\midrule
Disinformation Detection      & lr & $0.00096$ \\
           & Weight decay & $0.00018$ \\
           & Epochs & $50$ \\
           & Patience & $3$ \\
           & Seed & $[43, 434, 445]$ \\
           & Batch size & $128$ \\
\midrule
Infra & Hardware &a NVIDIA RTX 4090 (24 GB VRAM) GPU(1~2h for training and evaluate) \\
\bottomrule
\end{tabular}
\caption{Training hyper-parameters and configuration for Extax}
\label{tab:train_settings}
\end{table*}

\subsection{Baseline Implementation}
\label{app:baseline_imp}

Below, we report the detailed implementation of the reproduced baseline models. All baselines are evaluated under the same binary label space and dataset split protocol as ExTax. Specifically, CoAID and ECTF are used as source-domain datasets with training and validation splits, while CoAID, ECTF, EUDisinfo, ISOTFakeNews, and MultiDis are used for test evaluation. All labels are normalized into \texttt{real} and \texttt{fake}. For trainable baselines, hyperparameters are selected using the validation split. For LLM-based baselines, the source-domain training split is used as the support pool when retrieval, counterfactual generation, calibration, or knowledge construction is required. For each reproduced method, we retain the original high-level algorithmic design whenever possible and explicitly document the implementation-aligned adaptations required for our unified evaluation setting, including dataset loading, input/output formatting, label normalization, computational approximations, and unavailable-component substitutions.

\begin{itemize}
\item \textbf{RoBERTa} \citep{liuRoBERTaRobustlyOptimized2019}: For the baseline implementation, we utilize the \texttt{roberta-large} variant as the backbone encoder network. Given an input sequence, the token-level contextualized embeddings are extracted from the encoder, and a mean-pooling operation is subsequently applied across the sequence length to derive a unified sequence representation. This pooled representation is then fed into a sequence-level classification head structured as a multi-layer perceptron (MLP). Specifically, the classification head comprises a linear projection layer mapping the hidden size into a $256$-dimensional intermediate space, followed by a Gaussian Error Linear Unit (\text{GELU}) activation function and Layer Normalization. To prevent overfitting, a Dropout layer with a regularization rate of $0.25$ is applied before the final linear layer, which maps the features directly into the $2$-dimensional classification target space.

\item \textbf{DRES} \citep{Farhangian2025DRESFN}: 
We reproduce DRES using the provided Dynamic Representation and Ensemble Selection implementation and adapt its input/output interface to our unified data splits. The training set is constructed by merging the CoAID and ECTF training splits, and the validation set is constructed by merging the corresponding validation splits. The test set contains CoAID, ECTF, EUDisinfo, ISOTFakeNews, and MultiDis. Following the dynamic representation selection principle of DRES, we construct multiple TF-IDF feature spaces with maximum dimensions of $100$, $200$, $300$, $500$, $700$, and $1000$, respectively, using English stop-word removal. For each representation, we train a pool of five classifiers, including Logistic Regression, Bernoulli Naive Bayes, $k$-Nearest Neighbors, Random Forest, and Support Vector Machine. Hyperparameters are selected according to validation-set performance: Logistic Regression searches $C \in \{0.1,1,10\}$ with maximum iteration $2000$; Bernoulli Naive Bayes searches $\alpha \in \{0.5,1.0\}$; $k$-Nearest Neighbors searches $k \in \{3,5,7\}$; Random Forest uses $100$ estimators and searches maximum depth in $\{5,10,20\}$; and SVM searches $C \in \{0.1,1,10\}$ with an RBF kernel. Instance hardness is estimated by the $k$-Disagreeing Neighbors criterion with $k=5$. For each test instance, the representation with the lowest estimated hardness is selected dynamically. Dynamic ensemble selection is then performed using KNORA-Eliminate (KNORA-E; implemented as \texttt{KNORAE} in \texttt{deslib2}), with the region of competence size set to $5$. Following the local-oracle criterion of KNORA-E, only base classifiers that correctly classify all samples in the local region of competence are selected; if no classifier satisfies this criterion, the region is iteratively reduced and re-evaluated. The final prediction is obtained by majority voting among the classifiers selected by KNORA-E. The experiment is repeated with three random seeds, $\{43,434,445\}$.

\item \textbf{MSynFD} \citep{Xiao2024MSynFDMS}: 
We reproduce MSynFD as a syntax-aware neural fake news detection baseline based on the released implementation structure. The model uses \texttt{bert-base-uncased} as the textual encoder, followed by a Semantic Transformer module, a Graphformer-based multi-hop syntactic aggregation module, and a keyword debiasing branch. The hidden dimension is set to $768$, the number of attention heads is $12$, the dropout rate is $0.1$, and the hop number is set to $3$. The maximum input sequence length is $200$, and the maximum keyword sequence length is $50$. In our reproduced implementation, the syntactic adjacency matrix is approximated by a local dependency-distance window with window size $3$, rather than by running an external dependency parser. The final logits are computed by combining the main prediction branch and the keyword debiasing branch with weights $0.9$ and $0.1$, respectively. The training objective is the sum of the cross-entropy loss on the final prediction and $0.1$ times the cross-entropy loss on the keyword debiasing output. The model is trained for $3$ epochs with batch size $8$ and learning rate $1\times10^{-5}$ using AdamW. Following the actual optimizer parameter grouping in the implementation, non-bias and non-LayerNorm parameters use weight decay $0.01$, while bias and LayerNorm parameters use no weight decay. For each run, up to $2000$ training samples and $500$ validation samples are sampled using the corresponding random seed. Evaluation is performed with batch size $16$. We repeat the experiment with seeds $\{43,434,445\}$.

\item \textbf{RAEmoLLM} \citep{liuRAEmoLLMRetrievalAugmented2025}: 
We implement RAEmoLLM as an emotion-aware retrieval-augmented LLM baseline. The support pool is built from the CoAID and ECTF training splits. For each test instance, the model extracts affective information and retrieves source-domain examples as in-context demonstrations. The reproduced implementation uses keyword-based affective features and, when available, transformer-based emotion or sentiment models. For retrieval, semantic similarity is computed with \texttt{all-MiniLM-L6-v2} when the encoder is available, and affective similarity is computed from the extracted emotion-oriented representation. The retrieval score combines semantic and affective similarity with weights $0.45$ and $0.55$, respectively. If neural sentence encoders are unavailable, the implementation falls back to a TF-IDF retriever with unigram and bigram features. The retrieval size is set to top-$K=1$. The retrieved example, affective analysis, and target news text are inserted into a JSON-constrained prompt, and the LLM is required to output the predicted label, confidence score, and reasoning. The generated label is then normalized into the unified binary label space before evaluation.

\item \textbf{FACTUAL} \citep{liMitigatingBiasesLarge2025}: 
We reproduce FACTUAL as a counterfactual augmented calibration baseline and adapt it to binary fake news detection. The implementation first obtains an original LLM prediction and two debiased predictions, where the debiased prompts explicitly reduce sentiment-driven and target-prior-driven shortcuts. It then generates two counterfactual variants of the input: a causal counterfactual that minimally changes the text to flip the authenticity judgment, and a non-causal counterfactual that paraphrases the input while preserving the original stance. These counterfactual texts are re-evaluated by the LLM to estimate counterfactual fake probabilities. The final prediction probability is calibrated using the original prediction, the two debiased predictions, the counterfactual predictions, and an explicit bias signal. The calibration strength is set to $\alpha=0.35$. To avoid using target-test labels for calibration, the lightweight Logistic Regression calibration module is treated as an optional component and is used only when a calibrator trained on source-domain or validation data is explicitly available. Otherwise, we use the heuristic calibration rule implemented in the reproduced code. The final calibrated probability is thresholded at $0.5$ to obtain the binary prediction.

\item \textbf{DYNAMO} \citep{jinDynamicKnowledgeUpdateDriven2025}: 
We implement DYNAMO as a dynamic knowledge-update-driven LLM baseline with knowledge graph verification and Monte Carlo Tree Search. The source-domain support pool is built from the CoAID and ECTF training splits. The initial knowledge graph is constructed from reliable source-domain records labeled as \texttt{real}; if the number of reliable records is insufficient, the implementation falls back to the full support pool. In the reproduced setting, up to $10$ support records are used for initial knowledge graph construction, and at most $2$ entity-centered triples and $2$ event-centered triples are extracted from each selected text. For each test instance, DYNAMO decomposes the input into verifiable subclaims through Monte Carlo Tree Search. The reproduced MCTS configuration uses maximum depth $2$, maximum branching factor $2$, one simulation, and UCB1 exploration coefficient $1.414$. Each subclaim is verified against the constructed knowledge graph and assigned one of three verdicts: \texttt{supported}, \texttt{contradicted}, or \texttt{unverifiable}. The verification statistics and reasoning traces are then inserted into a JSON-constrained prompt, from which the LLM outputs the final binary label, confidence score, and reasoning. For efficiency, online knowledge updates are deferred during batch evaluation.
\begin{table*}[t]
\centering
\small
\begin{tabularx}{0.88\textwidth}{X}
\toprule
\rowcolor{gray!20} \textbf{Text}: \#Hope \#NYC \#NY \#USA \#World \#Health vigilance a \#vaccine is a year away \#Pandemic harmful is misinformation side effects~discrimination before \#Ebola \#SARS \#MERS there was \#AIDS a community marginalize...  \\
\rowcolor{gray!20} \textbf{Label}: Disinformation \quad \textbf{Genre}: Post  \\
\midrule
\textbf{Detected Taxonomies Type}: \newline
- Persuasion: Justification, Call, Manipulative wording.\newline
- Emotion: Hope, Sadness.\newline
- Narrative Roles: Deceptive Subversives, Marginalized Sufferers.\\
\bottomrule
\end{tabularx}
\caption{Case study 1.}
\label{tab:case1}
\end{table*}
\begin{table*}[t]
\centering
\small
\begin{tabularx}{0.88\textwidth}{X}
\toprule
\rowcolor{gray!20} \textbf{Text}: \underline{Title}: Trump Attempts New Year's Eve Do-Over Tweet And Gets CRUSHED By Americans He Called 'Enemies'
\underline{Content}: If Donald Trump had sent proper New Year's greetings to the American people to begin with, he wouldn't be getting humiliated. But he royally f*cked up on New Year's Eve morning by referring to millions...  \\
\rowcolor{gray!20} \textbf{Label}: Disinformation \quad \textbf{Genre}: News Article  \\
\midrule
\textbf{Detected Taxonomies Type}: \newline
- Persuasion: Attack on reputation, Justification, Simplification, Manipulative wording.\newline
- Emotion: Anger.\newline
- Narrative Roles: Overt Aggressors, Deceptive Subversives, Institutional Toxins.\\
\bottomrule
\end{tabularx}
\caption{Case study 2.}
\label{tab:case2}
\end{table*}
\item \textbf{PCoT} \citep{modzelewskiPCoTPersuasionAugmentedChain2025}: We implemented the baseline using the official code released by PCoT at \url{https://github.com/ArkadiusDS/PCoT}. During evaluation, we ran all three prompting strategies, namely VaN, Z-CoT, and DeF-SpeC, and selected the best-performing strategy, DeF-SpeC, as the final reported result.

\end{itemize}

For the three LLM-based baselines, RAEmoLLM, FACTUAL, and DYNAMO, we use the same OpenAI-compatible chat-completion backend with the configured model name \texttt{gpt-5-mini}. The decoding temperature is set to $0.1$, the maximum generation length is $256$ tokens, and the request timeout is set to $30$ seconds. Each LLM-based baseline is evaluated over $5$ repeated inference runs with fixed base seed $42$ for local sampling and preprocessing when applicable; variation across runs mainly comes from non-zero-temperature LLM generation. All LLM outputs are parsed through JSON-constrained generation and then normalized into the same binary label space as ExTax.

\subsection{Interpretability}
\label{app:interpretability}
\subsection{Global interpretability analysis}
This appendix provides additional analysis of the interpretability results reported in the main text. The global distribution of the three taxonomic dimensions---persuasion strategies, emotional manipulation, and narrative roles---is shown in Figure~\ref{fig:tax_distribution}. As illustrated, disinformation and real news exhibit markedly distinct structures within the constructed 17-dimensional taxonomic attribute space, providing intuitive evidence of the latent manipulative intents behind deceptive texts.

Specifically, concerning persuasion strategies, fake news relies heavily on ``Attack on Reputation'', ``Simplification'', and ``Manipulative Wording'', reflecting a typical discourse pattern that incites audiences by denigrating opposing targets through extreme language or misleads readers by omitting crucial information. This directly corroborates the findings in PCoT \citep{modzelewskiPCoTPersuasionAugmentedChain2025}. Regarding the emotional dimension, disinformation displays skewed distributions toward ``Anger'' and ``Fear''---two high-arousal negative emotions---which closely aligns with psychological theories \citep{jamiesonFlaggingEmotionalManipulation2025, zhaoHowIndividualsCope2024} suggesting that intense emotional stimuli accelerate the dissemination of misleading content. Furthermore, in shaping narrative roles, disinformation strongly tends to frame narrative subjects as negative archetypes, such as ``Deceptive Subversives'' and ``Institutional Toxins'', whereas real news leans more toward positive or constructive roles such as ``Ethical Stabilizers''.

To further elucidate how taxonomic attributes across different dimensions jointly manifest within texts, the Sankey diagram in Figure~\ref{fig:sankey_diagram} reveals their strong co-occurrence relationships. By tracing the convergence and divergence of the data flows, we observe several typical attribute binding patterns:
\begin{itemize}
\item \textit{High-frequency co-occurrence of aggressive persuasion and negative emotions.} From the left to the middle of the diagram, the persuasion strategy of ``Attack on Reputation'' exhibits a strong binding relationship with ``Anger'' and ``Fear''. This indicates that when disinformation employs aggressive discourse, it is frequently accompanied by intense, high-arousal negative emotional stimuli; the two act as closely coupled co-occurrence components within inciting texts.

\item \textit{Mutual corroboration of rational defense and constructive roles.} In contrast to the extreme co-occurrence patterns of disinformation, ``Justification''---a more neutral and objective persuasion strategy---has a high-probability co-occurrence with the ``None'' category in the emotional dimension. Texts without extreme emotional manipulation are also consistently associated with the positive narrative role of ``Ethical Stabilizers'', forming a core feature cluster of authentic news.

\item \textit{Chain transmission of non-manipulative characteristics.} When explicit manipulative intent is absent at one level, i.e., labeled as ``None'', this absence is likely to co-occur in subsequent levels. For instance, a large flow originating from ``None'' in persuasion strategies transitions into ``None'' in emotions and subsequently into ``None'' for narrative roles. This suggests that authentic news maintains a consistent non-manipulative pattern across multiple cognitive dimensions.
\end{itemize}

Overall, these distributional disparities show that ExTax not only improves the transparency of neural disinformation detection, but also captures multidimensional manipulative intents hidden within fluent texts, offering explanations that resemble the analytical process of human fact-checkers.

\subsubsection{Case Study Analysis}
\label{app:case_study_analysis}

To further demonstrate the instance-level interpretability of ExTax, we present two representative case studies in Table~\ref{tab:case1} and Table~\ref{tab:case2}. These examples show how taxonomy-aligned predictions can be translated into human-auditable explanations by identifying the persuasive strategies, emotional appeals, and narrative-role constructions that support each disinformation prediction.

\paragraph{Case Study 1: Health-related social media post.}
Table~\ref{tab:case1} presents a short pandemic- and vaccine-related social media post. Although the text is fragmented and hashtag-driven, ExTax identifies a coherent manipulation profile. In the persuasion dimension, the detected categories include \textit{Justification}, \textit{Call}, and \textit{Manipulative Wording}, indicating that the post attempts to legitimize its stance through emotionally charged public-health references while encouraging readers to adopt a particular viewpoint. Emotionally, ExTax detects both \textit{Hope} and \textit{Sadness}: vaccine-related references evoke hopeful expectations, whereas mentions of marginalized communities and severe disease histories trigger vulnerability and sympathy. In terms of narrative roles, \textit{Deceptive Subversives} and \textit{Marginalized Sufferers} suggest that the post frames some actors as potentially misleading while positioning affected communities as vulnerable victims. This case shows that ExTax can extract interpretable manipulation signals even from short and noisy social media posts.

\paragraph{Case Study 2: Political news article.}
Table~\ref{tab:case2} shows a long-form political article about Donald Trump's New Year's Eve tweet. Compared with the first case, this example contains more explicit evaluative and adversarial language. ExTax detects \textit{Attack on Reputation}, \textit{Justification}, \textit{Simplification}, and \textit{Manipulative Wording}, suggesting that the article relies on reputational attacks and reduces a complex political communication event into a morally one-sided conflict. The detected emotion, \textit{Anger}, is consistent with the article's confrontational and mocking tone. The narrative-role predictions, including \textit{Overt Aggressors}, \textit{Deceptive Subversives}, and \textit{Institutional Toxins}, further indicate that political actors are framed as visible antagonists and sources of institutional or social harm. Thus, the disinformation signal arises not from a single lexical cue, but from the alignment of aggressive persuasion, anger-based emotional manipulation, and negative role assignment.

\paragraph{Comparison between the two cases.}
Together, the two cases highlight the genre-adaptive interpretability of ExTax. The first case represents an implicit, compressed, and hashtag-driven manipulation pattern in short-form social media, while the second exhibits a more explicit pattern based on reputational attacks, anger, and antagonistic role construction in long-form political news. Despite these genre differences, both cases are mapped into the same 17-dimensional taxonomic space, enabling comparable explanations across posts and articles. These examples show that ExTax goes beyond binary classification by exposing how deceptive texts manipulate readers through rhetoric, emotion, and narrative framing.

\end{document}